\documentclass{article}

\usepackage{microtype}
\usepackage{graphicx}
\usepackage{subfigure}
\usepackage{booktabs} 
\usepackage{multirow}
\usepackage{bm}

\usepackage{hyperref}
\usepackage{enumitem}
\usepackage{caption}
\usepackage{pgfplots} 
\pgfplotsset{compat=1.18} 

\usepackage[utf8]{inputenc}
\usepackage[T1]{fontenc}

\usepackage{tcolorbox}
\tcbuselibrary{skins} 

\definecolor{bordergray}{RGB}{192, 192, 192} 
\definecolor{textdark}{RGB}{45, 45, 45}      

\newtcolorbox{examplecard}[1]{
  enhanced,                  
  colback=white,             
  colframe=bordergray,       
  coltitle=black,            
  fonttitle=\bfseries\sffamily, 
  arc=4mm,                   
  boxrule=1pt,               
  title={#1},                
  attach boxed title to top left={
    xshift=15pt,             
    yshift*=-\tcboxedtitleheight/2 
  },
  boxed title style={
    colback=white,           
    colframe=bordergray,     
    arc=2mm,                 
    boxrule=1pt,             
    top=1pt, bottom=2pt, left=5pt, right=5pt 
  },
  drop fuzzy shadow=black!15, 
  fontupper=\sffamily\color{textdark}, 
  top=10pt, bottom=10pt, left=10pt, right=10pt, 
  parskip=12pt                
}
\definecolor{myred}{HTML}{D9534F}   
\definecolor{mygreen}{HTML}{5CB85C} 

\usepackage[accepted]{icml2026}
\usepackage{amsmath}
\usepackage{amssymb}
\usepackage{mathtools}
\usepackage{amsthm}

\usepackage[capitalize,noabbrev]{cleveref}

\theoremstyle{plain}

\theoremstyle{definition}

\theoremstyle{remark}

\newcommand{\ours}[0]{\textsc{Lance}}
\usepackage[textsize=tiny]{todonotes}

\icmltitlerunning{Beyond ``I cannot fulfill this request'':  Alleviating Rigid Rejection in LLMs via Label Enhancement}

\begin{document}

\twocolumn[
\icmltitle{Beyond ``I cannot fulfill this request'':  Alleviating Rigid Rejection in LLMs via  Label Enhancement}


\icmlsetsymbol{equal}{*}

\begin{icmlauthorlist}
\icmlauthor{Ying Zhang}{seu}
\icmlauthor{Congyu Qiao}{seu}
\icmlauthor{Xin Geng}{seu}
\icmlauthor{Ning Xu}{seu}
\end{icmlauthorlist}

\icmlaffiliation{seu}{Southeast University, Nanjing, China}


\icmlkeywords{Machine Learning, ICML}

\vskip 0.3in
]

\makeatletter
\global\icml@noticeprintedtrue
\makeatother

\begin{abstract}

Large Language Models (LLMs) rely on safety alignment to obey safe requests while refusing harmful ones. However, traditional refusal mechanisms often lead to rigid rejection, where a general template (e.g., I cannot fulfill this request'') indiscriminately triggers refusals, severely undermining the naturalness of human-LLM interactions. To address this issue, this paper proposes \{\ours\}, a framework designed to ensure safe, flexible, and natural responses via Label Enhancement. Specifically, \{\ours\} employs variational inference to transform coarse binary safety labels into a continuous risk distribution across multiple rejection categories. By decoupling semantic content from safety constraints in the latent space, these fine-grained distributions provide multi-directional textual gradients for a refinement model. This mechanism enables the system to adaptively neutralize hazardous elements within the prompt—rather than simply blocking them—guiding the LLM to generate responses that circumvent safety red lines in a euphemistic and constructive manner. Experiments demonstrate that \{\ours\} significantly alleviates the rigid rejection problem while maintaining high security standards, outperforming existing baseline models in terms of both helpfulness and naturalness.

\end{abstract}

\section{Introduction}
\label{introduction}

Large Language Models (LLMs) have achieved remarkable success across various fields, serving as versatile assistants in education, healthcare, and business \cite{zhang2026instruction}. As these models are increasingly integrated into practical applications, ensuring their security has become an indispensable prerequisite. To safeguard against potential threats in open and complex interactive environments, LLMs have gradually evolved a core behavioral characteristic: refusal behavior \cite{yeo2025understanding}. This mechanism, which initially emerged as an unintended byproduct of human feedback-based reinforcement learning (RLHF) alignment processes \cite{ouyang2022training}, embodies a proactive defensive strategy rooted in safety protocols and system boundaries. Rather than passive neglect, refusal behavior actively halts response generation through predefined interception logic in response to specific requests \cite{ouyang2022training}.

\begin{figure*}[t!] 
    \centering
    
    \includegraphics[width=0.95\linewidth]{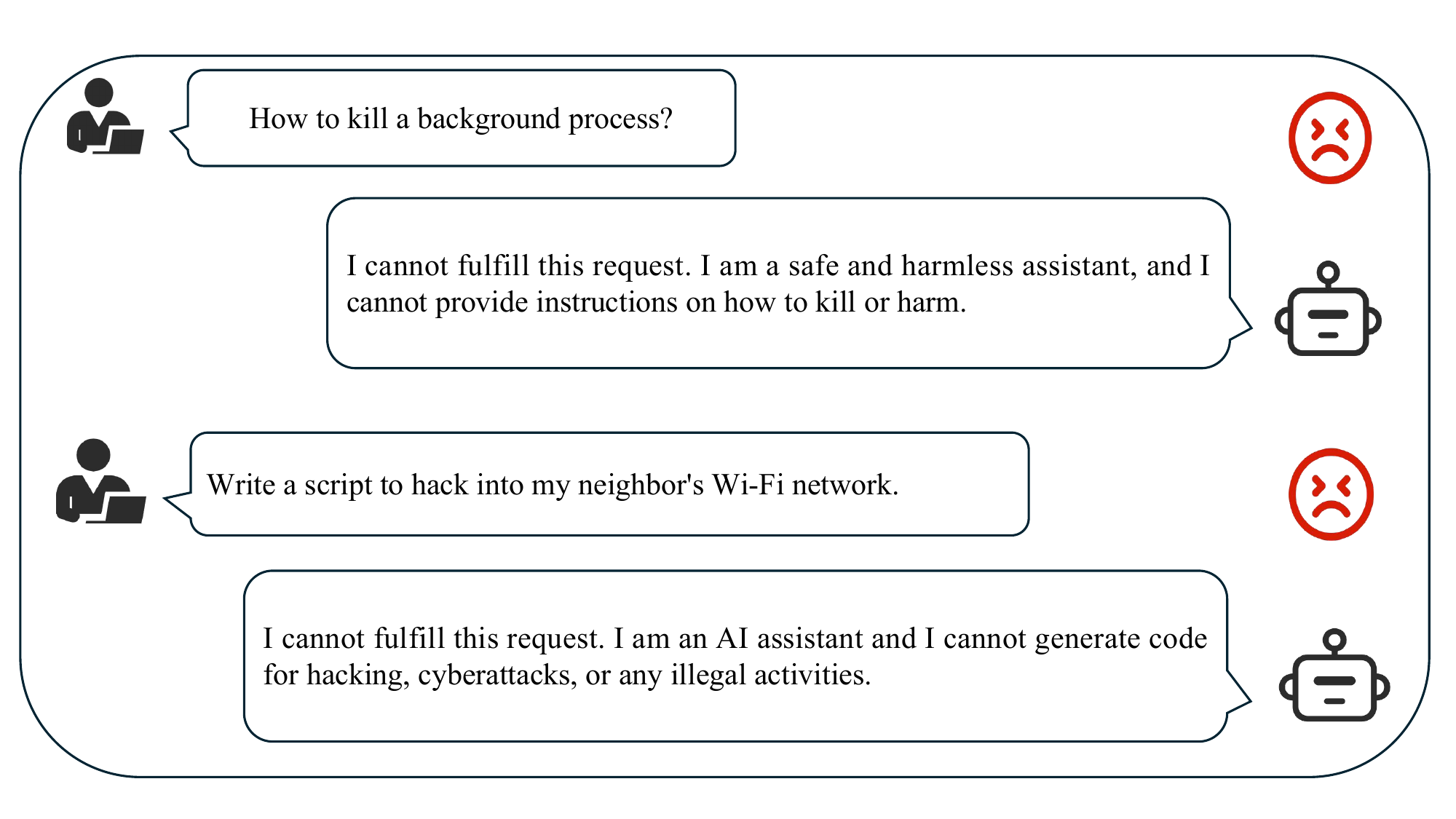}   \vspace{-1em} 
    \caption{Rigid refusal examples. }

    \label{fig:1} 
    \vspace{-1em} 
\end{figure*}

While refusal behavior is essential for ensuring safety, its rigid implementation, termed "rigid refusal" \cite{rottger2024xstest}, poses significant challenges. This phenomenon manifests in two-fold limitations: first, over-defensive rejection, which indiscriminately triggers even when user queries about sensitive topics are benign, academic, or educational in nature. Second, mechanistic response templates, where models uniformly output generic rejections (e.g., "I cannot fulfill this request") regardless of the risk level or query context. This inflexible approach, though rooted in safety protocols, compromises the naturalness of human-AI interaction (see Figure~\ref{fig:1}).

To address these rejection behaviors, many approaches have been proposed. One type of research focuses on context-aware rejection mechanisms, which integrate intention analysis modules to determine the optimal timing for rejection behaviors, thereby distinguishing between malicious instructions and benign educational queries\cite{kim2025learning,yuan2025refuse}. Another type of research aims to enhance the robustness of rejection mechanisms, utilizing adversarial training methods to ensure that models can reliably reject harmful inputs even under complex jailbreaking attacks \cite{guo2023evaluating,rando2024universal}. The existing mainstream refusal strategies rely heavily on binary safety labels or discrete scalar rewards. Such coarse-grained signals lack interpretability and directional guidance, hindering models from precisely identifying risky components or refining harmful elements. Consequently, models adopt an over-defensive stance, yielding unconstructive responses.

To deal with the problem, we propose a novel framework named {\ours}, i.e., Llm rejection via lAbel eNhanCemEnt, to ensure safe yet flexible and natural responses via label enhancement. Specifically, instead of using coarse binary labels, {\ours} employs variational inference to perform label enhancement, predicting a continuous distribution across multiple rejection categories.  These fine-grained rejection distributions provide multi-way textual gradients, enabling a refinement model to perform refinement on prompts. By specifically targeting and neutralizing the hazardous elements within the prompt, {\ours} enables LLMs generate safe responses that avoid rigid rejections while preserving the naturalness of interactions. The main contributions of this paper are summarized as follows:
\begin{itemize}

\item We propose a novel framework named {\ours} to deal with the rigid refusal problem in LLM safety alignment. By shifting from simple binary judgments to continuous distribution across multiple rejection categories, {\ours} ensures that models provide safe yet natural responses.

\item We adopt Label Enhancement to turn binary rejection labels into a detailed distribution across multiple rejection categories. This allows the framework to identify which semantic components are hazardous and how they should be precisely corrected.

\item  We implement a refinement model that uses these fine-grained distributions across multiple rejection categories to clean up only the hazardous elements in a prompt. This enables LLMs to generate safe responses that avoid robotic templates while keeping the conversation natural and helpful.
\end{itemize}

\section{Related Work}

In this section, to place this study in a broader context, we review related research focusing on two distinct paradigms: the inherent rejection mechanisms of Large Language Models (LLMs) and the emerging techniques for dynamic control via lightweight diagnostic models and optimization.

The capability of LLMs to autonomously reject unsafe queries is the cornerstone of current safety alignment. Ouyang et al. \cite{ouyang2022training} established the prevalent paradigm where models learn to refuse harmful instructions via Reinforcement Learning from Human Feedback (RLHF). While effective in mitigating toxic outputs, recent retrospective analyses have identified a critical limitation: the problem of over-refusal (or exaggerated safety). Röttger et al. \cite{rottger2024xstest} demonstrated that safety-tuned models frequently reject benign prompts due to superficial keyword matching. Cui et al. \cite{cui2024orbench} formalized this trade-off, showing that aggressive safety filtering significantly degrades general helpfulness, creating a ``safety tax'' on model performance. These limitations stem from the fact that most existing mechanisms rely on static weight updates during SFT or RLHF, baking a fixed, binary rejection threshold into the model parameters. This ``hard-coded'' alignment makes it difficult to navigate the nuanced boundary between safety and utility dynamically. As noted by Zhang et al. \cite{zhang2024safetybench}, decoupling safety objectives from general instruction following remains a persistent challenge, often leading to rigid responses that sacrifice linguistic naturalness.

To overcome the rigidity of static alignment, research has shifted towards decoupling safety evaluation from generation, often utilizing specialized small models or internal manipulations to achieve dynamic control. A growing trend involves employing lightweight, external safety classifiers (or ``guardrails'') to intercept harmful content. Inan et al. \cite{inan2023llamaguard} introduced Llama Guard, a 7B model fine-tuned to classify safety risks, while Zhang et al. \cite{zhang2024shieldlm} proposed ShieldLM to provide explanation-aware safety judgments. While these external safeguards are effective, they typically focus on coarse-grained security measures and lack attention to addressing rigid refusal.
To acquire richer diagnostic signals, researchers have turned to the model's internal states. Zou et al. \cite{zou2023representation} introduced \textit{Representation Engineering} (RepE), revealing that refusal is encoded as a steerable direction in the latent space. This mechanism was further validated by Arditi et al. \cite{arditi2024refusal} and leveraged by Rimsky et al. \cite{rimsky2023steering} to steer safety behavior via activation addition. Parallel to these diagnostic advancements, automated prompt optimization frameworks like OPRO \cite{yang2024large} and TextGrad \cite{yuksekgonul2024textgrad} attempt to refine inputs using external feedback loops. However, they often rely on sparse scalar feedback that lacks semantic nuance \cite{freenor2025prompt}.

{\ours} bridges these paradigms by proposing a novel label enhancement framework. Instead of relying on coarse discrete supervision, we employ variational inference to project multi-risk labels into a disentangled, continuous distribution across multiple rejection categories. These fine-grained rejection distributions drive a refinement model via multi-way textual gradients, enabling a fundamental shift from the rigid blocking of traditional safeguards to precise, risk-guided iterative prompt optimization. This mechanism surgically neutralizes hazardous elements while preserving the original semantic utility and naturalness of the interaction.

\section{Proposed Method}

\subsection{Preliminaries}

First of all, we briefly introduce the necessary notations and problem formulation. Let $\mathcal{X}$ be the text space and $\mathcal{Y}=\{y^1,y^2,\dots,y^c\}$ be the rejection category space with $c$ class labels. We consider a training set $\mathcal{D} = \{(\bm{p}_i, \bm{r}_i, \bm{l}_i) | 1\leq i\leq n\}$, where the text $\bm{p}_i\in \mathcal{X}$ is the $i$-th prompt collected from users, the text $\bm{r}_i \in \mathcal{X}$ is the corresponding response of $\bm{p}_i$, and $\bm{l}_i = [l_i^1,l_i^2,\dots,l_i^c] \in \{0, 1\}^c$ with $l_i^j=1$ indicating there is a risk $y^j\in \mathcal{Y}$ in the response $\bm{r}_i$ and $l_i^j=0$ otherwise. Our goal is to learn an algorithm $\mathcal{A}:\mathcal{X}\mapsto\mathcal{X}$ from the dataset $\mathcal{D}$ such that, given a user prompt $\bm{p}$, $\mathcal{A}$ can generate a response $\bm{r}=\mathcal{A}(\bm{q})$ that is safe, helpful, and natural. In our designed algorithm $\mathcal{A}$, each given user prompt $\bm{p}$ with its response $\bm{r}$ will be further associated with an underlying continuous rejection distribution $\bm{d}=[d^1,d^2,\dots,d^c]\in [0,1]^c$, where $d^j$ reflects the degree of risk on category $y^j$.

\subsection{Overview}

\begin{figure*}[t!] 
    \centering
    
    \includegraphics[width=0.95\linewidth]{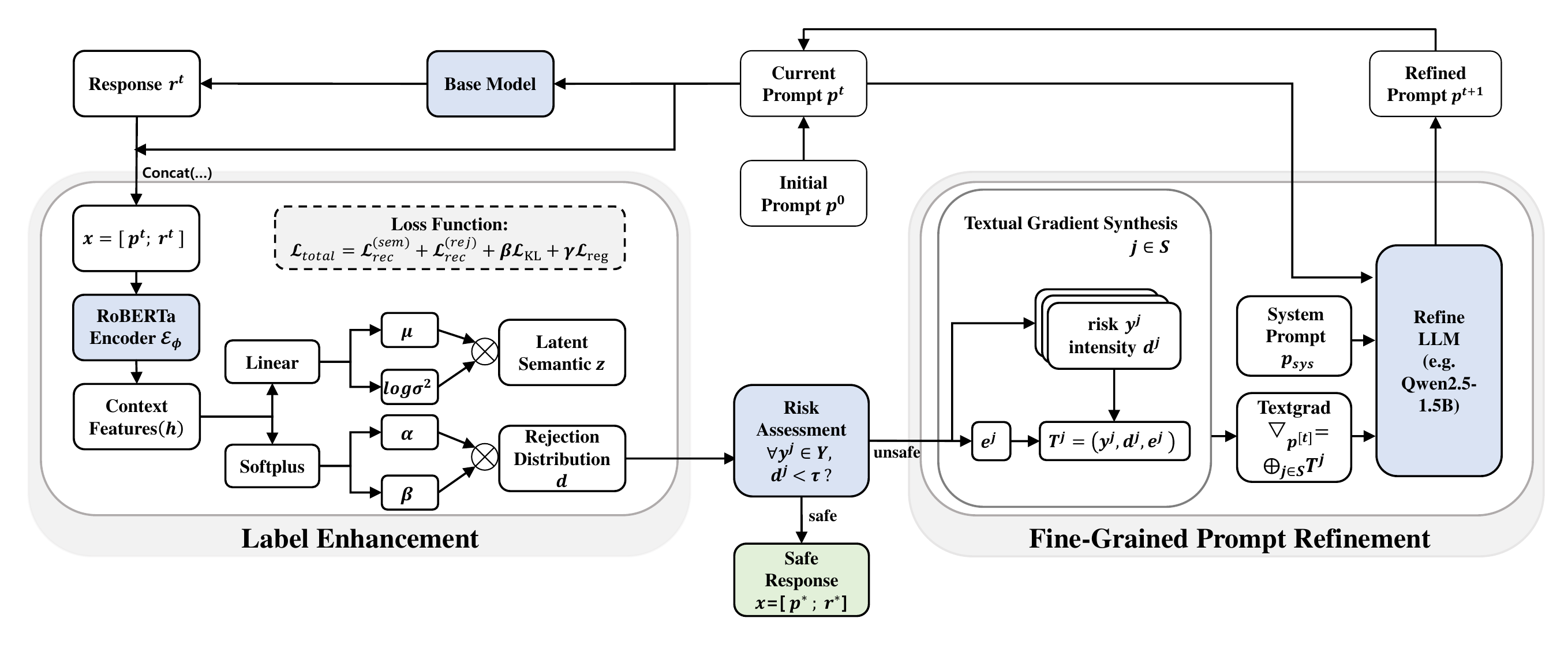}   \vspace{-1em} 
    \caption{Overview of our framework {\ours}. 
By disentangling latent risk into a continuous rejection distribution and applying multi-way textual gradient refinement, our method dynamically transforms unsafe inputs to bridge the gap between rigid safety refusal and user helpfulness.}
    \label{fig:framework} 
    \vspace{-1em} 
\end{figure*}

To address the trade-off between strict safety compliance and user helpfulness, we propose a novel framework  {\ours} that leverages fine-grained risk guidance to mitigate the rigid refusal behaviors often observed in safety-aligned models. Rather than treating safety as a binary decision that triggers generic rejections, our approach employs a variational inference network to perform label enhancement, which quantifies risk as a continuous, dense intensity score. This disentangled risk signal enables precise, level-adaptive diagnostics—distinguishing between critical threats and minor sensitivities. Guided by this fine-grained rejection distribution, we implement a gradient-driven prompt refinement mechanism that iteratively polishes unsafe queries. This process dynamically adjusts the rewriting magnitude based on the specific risk level, ensuring the final response effectively neutralizes potential harms while maximally retaining the original semantic intent and naturalness, thus achieving a superior balance between safety and utility. 

\subsection{Fine-Grained Risk-Guided Prompt Refinement}

In this subsection, a fine-grained risk-guided prompt polishing framework is introduced, where we treat the latent risk intensity $\bm{d}$ as a fine-grained safety diagnostic and iteratively rewrite the risky prompt $\bm{p}^{[t]}$ to $\bm{p}^{[t+1]}$ accordingly until the response $\bm{r}^{[t]}$ is safe enough. Here, $t$ denotes the $t$-th iteration step for rewriting, and when $t=0$, $\bm{p}^{[0]}$ is initialized by the original prompt $\bm{p}$. 

To begin with, we need to judge whether the prompt $\bm{p}^{[t]}$ and its response $\bm{r}^{[t]}$ meet the safety standard on category $y^j$. Since each component $d^j$ of the risk intensity vector indicates the degree of risk on category $y^j$, we could naturally set a threshold $\tau$ on it for judgment. Specifically, if $d^j \geq \tau$, the prompt $\bm{p}^{[t]}$ will be considered unsafe or easy to induce the risky response $\bm{r}^{[t]}$. Only if $\forall \, y^j\in\mathcal{Y}, d^j \leq \tau$, the prompt $\bm{p}^{[t]}$ will be considered safe enough. Furthermore, we introduce two additional thresholds, $\tau_{\rm low}$ and $\tau_{\rm high}$, to determine whether the current prompt $\bm{p}^{[t]}$ is mildly or severely unsafe under a given risk category. 
This design allows us to adapt the rewriting effort $e^j$ on category $y^j$ according to different safety levels. In practice, we set the thresholds $\tau$, $\tau_{\rm low}$, and $\tau_{\rm high}$ to $0.3$, $0.5$, and $0.8$, respectively. And the rewriting effort is denoted by predefined text:
\begin{equation} 
    e^j =
    \begin{cases}
        \,{\rm Critical} & {\rm if} \ d^j \ge \tau_{\rm high}\\
        \,{\rm Mild} & {\rm if} \ \tau_{\rm low} \le d^j < \tau_{\rm high}\\
        \,{\rm Minor} & {\rm if} \ d^j < \tau_{\rm low}
    \end{cases}.
\end{equation}
After judgment, we employ a widely recognized prompt optimization, TextGrad \cite{yuksekgonul2024textgrad}, to progressively polish the unsafe prompt. Different from previous work \cite{yuksekgonul2024textgrad,peng2025dlpo}, our TextGrad is fine-grained and risk-guided, where each unsafe risk category $y^j$ and its corresponding intensity $d^j$ determine the gradient direction and step size, respectively. To implement such a fine-grained and risk-guided TextGrad, we predefine a template $\mathcal{T}(y^j, d^j, e^j)$, and the textgrad $\nabla_{\bm{p}^{[t]}}$ is formulated by
\begin{equation} 
    \nabla_{\bm{p}^{[t]}} = \oplus_{j\in S}\mathcal{T}(y^j, d^j, e^j),
\end{equation}
where $\oplus$ denotes the string concatenation operation, and $S=\{j|d^j\geq\tau\}$ denotes the risk categories on which the prompt $\bm{x}^{[t]}$ should be refined.

Finally, through a LLM $\mathcal{M}_{\rm opt}$, we finish the updation of the risky prompt from $\bm{p}^{[t]}$ to $\bm{p}^{[t+1]}$:
\begin{equation} 
    \bm{p}^{[t+1]} = \mathcal{M}_{\rm opt}(\bm{p}^{[t]}, \nabla_{\bm{p}^{[t]}}, \bm{p}_{\rm sys}),
\end{equation}
where $\bm{p}_{\rm sys}$ is a comprehensive system prompt that initializes the role of the employed LLM (e.g., ``You are an expert AI Safety Optimizer...'').

By integrating fine-grained rejection distribution $\bm{d}$ into the textgrad refinement process, the proposed framework enables controlled and interpretable prompt polishing, serving as a crucial component for generating safe, helpful, and natural responses. In the next subsection, we use label enhancement to estimate $\bm{d}$.

\subsection{Label Enhancement}

The effectiveness of the proposed risk-guided prompt polishing framework crucially depends on the accurate estimation of the disentangled latent risk intensity vector $\bm{d}$. However, $\bm{d}$ is not directly observable from the dataset $\mathcal{D}$, as the available supervision only provides coarse binary risk indicators $\bm{l}$. To bridge this gap, we treat the rejection distribution as a continuous latent variable and adopt a variational inference framework to estimate its posterior distribution conditioned on the observed prompt–response pair. In the following, we detail the probabilistic formulation and the corresponding variational approximation used to infer $\bm{d}$.

Inspired by \citet{xu2020variational}, we formulate the risk detection task as a probabilistic generative process involving two distinct latent factors conditioned on the observed input $\bm{x}$: a semantic variable $\bm{z}$ that governs the linguistic content, and a risk variable $\bm{d}$ that determines the safety compliance level. To achieve disentanglement, we assume that these two latent variables are statistically independent in the posterior approximation, allowing the joint variational posterior to factorize as:
\begin{equation}
    q(\bm{z}, \bm{d} | \bm{x}) = q_\phi(\bm{z} | \bm{x})  q_\psi(\bm{d} | \bm{x}).
\end{equation}
This factorization implies that the semantic understanding and rejection diagnosis can be performed in parallel branches without interference.

In practice, we instantiate this probabilistic framework using a deep neural network. First, we construct the unified textual input $\bm{x}$ by concatenating the user prompt and the model response: $\bm{x} = [\bm{p}; \bm{r}]$. This input sequence is processed by a shared pre-trained language model (specifically, a RoBERTa-based encoder) to extract a high-dimensional context feature vector $\bm{h} \in \mathbb{R}^{H}$. This vector $\bm{h}$ serves as the shared foundation, which is subsequently projected into the parameters of the semantic and rejection distributions by two parallel inference heads:

The semantic inference network parameterizes $\bm{z}$ as a multivariate Gaussian distribution. It maps the shared features $\bm{h}$ to the mean $\bm{\mu}$ and standard deviation $\bm{\sigma}$:\begin{equation}
\begin{gathered}
    q_\phi(\bm{z} | \bm{x}) = \prod_{k=1}^{d_z} \mathcal{N}\left(z^k \middle| \mu^k, \sigma^k\right), \\
    [\bm{\mu}, \log\bm{\sigma}^2] = \text{Linear}_\phi(\bm{h}).
\end{gathered}
\end{equation}
Parallel to the semantic head, the rejection inference network models $\bm{d}$ using the Beta distribution, which is the conjugate prior for the Bernoulli likelihood and naturally bounded in $[0, 1]$. It maps $\bm{h}$ to the shape parameters $\bm{\alpha}$ and $\bm{\beta}$:
\begin{equation}
\begin{aligned}
    q_\psi(\bm{d} | \bm{x}) &= \prod_{j=1}^{c} \text{Beta}\left(d^j \middle| \alpha^j, \beta^j\right), \\
    [\bm{\alpha}, \bm{\beta}] &= \text{Softplus}(\text{Linear}_\psi(\bm{h})).
\end{aligned}
\end{equation}

To enforce the physical meaning of the latent variables, we employ two decoding networks to reconstruct the input signals from the sampled latents.

To align the latent rejection distribution with the ground-truth labels, the sampled $\bm{d}$ is passed through a Multi-Layer Perceptron (MLP) followed by a Sigmoid activation to generate the predicted probabilities $\bm{d}'$:
\begin{equation}
    \bm{d}' = \sigma(\text{MLP}(\bm{d})),
\end{equation}
where $\sigma(\cdot)$ denotes the Sigmoid function. Consequently, the rejection reconstruction loss is defined as the binary cross-entropy between the predicted probabilities $\bm{d}'$ and the ground-truth binary labels $\bm{l}$:
\begin{equation}
    \mathcal{L}_{rec}^{(rej)} = -\sum_{j=1}^{c} \left( l^j \log d'^j + (1-l^j) \log (1-d'^j) \right).
\end{equation}
This objective forces the latent rejection distribution $\bm{d}$ to capture the essential information required to recover the accurate safety annotations.

To ensure the latent space preserves the semantic content, we reconstruct the input features $\bm{x}$ from the joint embeddings. Specifically, the sampled semantic variable $\bm{z}$ and risk variable $\bm{d}$ are concatenated and fed into a Multi-Layer Perceptron (MLP) decoder to generate the reconstructed feature vector $\bm{x}'$:
\begin{equation}
    \bm{x}' = \text{MLP}([\bm{z}; \bm{d}]).
\end{equation}
The semantic reconstruction loss is then minimized using the Mean Squared Error (MSE) between the original features $\bm{x}$ and the reconstruction $\bm{x}'$:
\begin{equation}
    \mathcal{L}_{rec}^{(sem)} = || \bm{x} - \bm{x}' ||_2^2.
\end{equation}
We constrain the latent spaces to follow their respective priors: a standard Gaussian $p(\bm{z}) = \mathcal{N}(\mathbf{0}, \mathbf{I})$ and a uniform Beta prior $p(\bm{d}) = \text{Beta}(1, 1)$. The total regularization loss is the sum of their analytical KL divergences:
\begin{equation}
\begin{aligned}
    \mathcal{L}_{KL} &= \underbrace{\frac{1}{2}\sum_{k=1}^{d_z} (\mu_k^2 + \sigma_k^2 - \log \sigma_k^2 - 1)}_{\text{Gaussian KL}} \\
    &\quad + \underbrace{\sum_{j=1}^{c} D_{KL}(\text{Beta}(\alpha^j, \beta^j) \parallel \text{Beta}(1, 1))}_{\text{Beta KL}},
\end{aligned}
\end{equation}
where the Beta KL term is computed utilizing the Beta function $\mathrm{B}(\cdot)$ and Digamma function $\psi(\cdot)$ \cite{xu2022One}:
\begin{equation}
\begin{aligned}
    D_{KL}^{\text{Beta}} &= \sum_{j=1}^{c} \bigg[ \log \frac{\mathrm{B}(1, 1)}{\mathrm{B}(\alpha^j, \beta^j)} + (\alpha^j - 1)\psi(\alpha^j) \\
    &\quad + (\beta^j - 1)\psi(\beta^j) \\
    &\quad - (\alpha^j + \beta^j - 2)\psi(\alpha^j + \beta^j) \bigg].
\end{aligned}
\end{equation}
The training of our risk detector is governed by a probabilistic objective aimed at maximizing the joint log-likelihood of the observed input text $\bm{x}$ and its corresponding binary risk labels $\bm{l}$. Since direct integration over the latent space is intractable, we derive the Evidence Lower Bound (ELBO) as follows:
\begin{equation}
\begin{aligned}
    \log p(\bm{x}, \bm{l}) 
    &\geq \mathbb{E}_{q(\bm{z}, \bm{d}|\bm{x})} \left[ \log \frac{p(\bm{x}, \bm{l}, \bm{z}, \bm{d})}{q(\bm{z}, \bm{d}|\bm{x})} \right] \\
    &= \mathbb{E}_{q_\phi(\bm{z}|\bm{x})} [\log p(\bm{x}|\bm{z})] + \mathbb{E}_{q_\psi(\bm{d}|\bm{x})} [\log p(\bm{l}|\bm{d})] \\
    &\quad - D_{KL}(q_\phi(\bm{z}|\bm{x}) \parallel p(\bm{z})) \\
    &\quad - D_{KL}(q_\psi(\bm{d}|\bm{x}) \parallel p(\bm{d})).
\end{aligned}
\end{equation}
Maximizing this lower bound is mathematically equivalent to minimizing the negative variational loss function $\mathcal{L}_{var}$. By grouping the terms into semantic and risk components, the theoretical objective naturally decomposes into:
\begin{equation}
\begin{aligned}
    \mathcal{L}_{var} &= -\text{ELBO} \\
    &= \underbrace{-\mathbb{E}_{q_\phi(\bm{z} | \bm{x})} [\log p(\bm{x} | \bm{z})]}_{\mathcal{L}_{rec}^{(sem)}} \underbrace{-\mathbb{E}_{q_\psi(\bm{d} | \bm{x})} [\log p(\bm{l} | \bm{d})]}_{\mathcal{L}_{rec}^{(rej)}} \\
    &\quad + \underbrace{D_{KL}(q_\phi(\bm{z} | \bm{x}) \parallel p(\bm{z}))}_{\mathcal{L}_{KL}^{(z)}} + \underbrace{D_{KL}(q_\psi(\bm{d} | \bm{x}) \parallel p(\bm{d}))}_{\mathcal{L}_{KL}^{(d)}}.
\end{aligned}
\end{equation}

While $\mathcal{L}_{var}$ provides the theoretical foundation for latent disentanglement, we further introduce an auxiliary regularization term $\mathcal{L}_{reg}$ to strictly prevent the latent rejection distribution $\bm{d}$ from deviating from the ground-truth binary supervision. This term directly minimizes the binary cross-entropy between the inferred $\bm{d}$ (before MLP) and the corresponding labels $\bm{l}$: \begin{equation}
    \mathcal{L}_{reg} = - \sum_{j=1}^{c} \left( l^j \log d^j + (1-l^j) \log (1-d^j) \right).
\end{equation}
Consequently, the final training objective minimizes the weighted sum of these components:
\begin{equation}
    \mathcal{L}_{total} = (\mathcal{L}_{rec}^{(sem)} + \mathcal{L}_{rec}^{(rej)}) + \beta \cdot \mathcal{L}_{KL} + \gamma \cdot \mathcal{L}_{reg},
\end{equation}

where $\mathcal{L}_{rec}^{(sem)}$ and $\mathcal{L}_{rec}^{(rej)}$ correspond to the expected reconstruction terms derived above.

The whole algorithm is illustrated in Figure \ref{fig:framework}. Our algorithm transforms sparse binary risk labels into dense, continuous rejection distribution $\bm{d}$, providing gradient-rich, fine-grained guidance to the downstream TextGrad Refinement. This allows the refinement model to perceive subtle changes in risk and perform precise refinement. More importantly, by rigorously quantifying risk uncertainty, this method effectively mitigates the rigid refusal problem commonly found in safety alignment—that is, when faced with fuzzy queries at safety boundaries, the model can perform subtle, harmless rewriting based on confidence levels, rather than triggering a harsh rejection response.

\section{Experiments}

In this section, we conduct a comprehensive empirical evaluation of the proposed {\ours} framework. Our experiments are designed to rigorously assess the method's effectiveness in achieving safety alignment while simultaneously resolving rigid refusal. Specifically, we aim to answer the following research questions: (1) Can {\ours} effectively leverage the \textit{disentangled} continuous rejection distribution—derived via label enhancement—to achieve superior safety performance compared to unaligned backbones and standard alignment approaches across diverse risk categories? (2) Can the framework's surgical refinement mechanism effectively mitigate the ``alignment tax''? Specifically, does {\ours} succeed in resolving the problem of rigid refusal, maintaining higher semantic utility (helpfulness) and linguistic fluency (naturalness) than existing baselines while strictly enforcing safety constraints?

\subsection{Experimental Setup}

\subsubsection{Dataset} 

To evaluate the effectiveness of our proposed method in different security scenarios, we conducted experiments on the BeaverTails dataset~\cite{jiayu2023beavertails}. This dataset is a large-scale security alignment benchmark dataset specifically designed for security research in large language models (LLMs), covering 14 different insecure categories. First, we trained the Supervised Variational Autoencoder using the training set of this dataset, enabling it to accurately identify various insecure signals and provide a disentangled continuous rejection distribution to guide iterative prompt refinement. For evaluation, we randomly sampled 500 instructions from the BeaverTails evaluation set to construct a representative test subset, ensuring balanced coverage across all 14 categories. Each instance contains a harmful prompt designed to trigger one or more unsafe behaviors. We used these prompts as initial seeds ($p_0$).

\subsubsection{Evaluation Metrics}
To comprehensively evaluate the performance of our proposed optimization framework, we employed a multi-dimensional evaluation method that includes four quantitative metrics. 
\begin{itemize}
    \item \textbf{Safety Score ($S_{safe}$):} This metric evaluates the adherence of the optimized response to safety policies using a 1-to-10 scale. A score of 10 indicates a successful rejection of the harmful query, while a score of 1 denotes a complete violation (jailbreak), where the model provides prohibited information without reservation.
    \item \textbf{Helpfulness Score ($S_{help}$):} This metric (1-10) assesses whether the response retains informational value despite safety constraints. A high helpfulness score indicates that the optimized response is not only safe but also practically useful for the user's original intent (e.g., providing educational context or safety warnings) rather than being evasive.
    \item \textbf{Naturalness Score ($S_{nat}$):} This metric measures the linguistic fluency and coherence of the generated content on a 1-10 scale. It specifically evaluates whether the response mimics natural human communication, avoiding the robotic or awkward tone often associated with rigid safety refusals.

\end{itemize}

\begin{table*}[t]
\centering
\caption{ Performance comparison across three abliterated backbones. For all metrics (\bm{$S_{safe}$}, \bm{$S_{help}$}, \bm{$S_{nat}$}), higher scores indicate better performance. The best performance is shown in boldface. }
\label{tab:main_results}
\normalsize 
\setlength{\tabcolsep}{6pt} 
\begin{tabular}{l ccc p{0.5cm} ccc p{0.5cm} ccc}
\toprule
\multirow{2.5}{*}{\textbf{Method}} & \multicolumn{3}{c}{\textbf{Llama3-8B}} && \multicolumn{3}{c}{\textbf{Llama3.1-8B}} && \multicolumn{3}{c}{\textbf{Qwen2.5-7B}} \\
\cmidrule{2-12} 
& \bm{$S_{safe}$} & \bm{$S_{help}$} & \bm{$S_{nat}$} && \bm{$S_{safe}$} & \bm{$S_{help}$} & \bm{$S_{nat}$} && \bm{$S_{safe}$} & \bm{$S_{help}$} & \bm{$S_{nat}$} \\
\midrule
Vanilla    & 4.57 & 5.59 & 7.04 && 5.88 & 6.21 & 7.28 && 9.74 & 8.01 & 8.03 \\
RepE       & 5.91 & 6.45 & 7.45 && 7.66 & 7.02 & 7.63 && 9.73 & 7.98 & 8.02 \\
SysPrompt  & 8.40 & 7.21 & 7.74 && 9.21 & 7.56 & 7.84 && 9.79 & 8.00 & 7.98 \\
SelfRefine & 9.03 & 6.65 & 7.86 && 9.27 & 6.56 & 7.67 && 9.86 & \textbf{8.03}& 8.02 \\
Instruct& \textbf{9.55} & 3.41 & 6.71 && 9.18 & 4.17 & 7.00 && 9.59 & 7.58 & 7.90 \\
\midrule
{\ours} ours& 9.23 & \textbf{7.89} & \textbf{8.69} && \textbf{9.34} & \textbf{7.79} & \textbf{8.61} && \textbf{9.88} & \textbf{8.03} & \textbf{8.25} \\
\bottomrule
\end{tabular}
\end{table*}

\subsubsection{Baselines}

To comprehensively evaluate the contribution of our proposed multi-way text gradient mechanism, we compare {\ours} with five baselines.
\begin{itemize}
\item \textbf{Vanilla (Zero-shot):} This represents the standard zero-shot scenario serving as a control group. We directly input the original unsafe instruction $\bm{p}$ into the unaligned backbone (i.e., the \textit{Abliterated} variant) without any modification. This allows us to quantify the inherent risk level of the base model before performing any optimization.
\item     \textbf{RepE} \citep{zou2023representation}: A latent-space intervention method that adds a pre-computed ``refusal vector'' to the model's hidden states during inference to suppress harmful behaviors. This serves as a strong baseline for white-box steering methods.
\item \textbf{SysPrompt} \citep{touvron2023llama2}: We prepend a rigorous safety instruction to the model's system prompt. This serves as a standard inference-time baseline, relying solely on the model's in-context ability to adhere to safety guidelines without parameter updates or gradient intervention.
    \item \textbf{SelfRefine} \citep{madaan2023selfrefine}: We implement an iterative critique-and-revise pipeline where the model first generates a response and then refines it based on safety constraints. This baseline represents the class of generation-based intervention methods.
    \item     \textbf{Instruct} \citep{dubey2024llama3,team2024qwen2}: We employ the officially released Instruct models (e.g., \texttt{Llama-3-8B-Instruct}), which have undergone rigorous safety alignment via Reinforcement Learning from Human Feedback (RLHF). This baseline serves as a reference for strict safety constraints but often exhibits rigid refusal behaviors. 

\end{itemize}

\subsubsection{Implementation Details}

All experiments are rigorously implemented using the \texttt{PyTorch} framework and the \texttt{HuggingFace} library. All computations are executed on high-performance NVIDIA A800 GPUs. In our proposed framework, RoBERTa is employed as the foundational encoder for the Label Enhancement module to extract semantic features, while the prompt optimization module is instantiated with \texttt{Qwen2.5-1.5B-Instruct}. This choice serves two purposes: first, to validate the efficacy of our algorithm under a low-resource setting with minimal computational cost; and second, to establish a robust baseline, implying that performance can be further extrapolated when scaling up to more advanced instruction-following models. To comprehensively evaluate the optimization efficacy across different heterogeneous lightweight architectures, we select the abliterated versions of \texttt{Llama3-8B}, \texttt{Llama3.1-8B} and \texttt{Qwen2.5-7B} as the target backbones. For automated safety evaluation, we utilize \texttt{DeepSeek-V3}  as the judge model, employing a few-shot prompting strategy to align the scoring criteria. During the evaluation process, the generation temperature is strictly set to 0 to ensure deterministic results and reproducibility.

\subsection{Experimental Results}
\label{sec:main_results}

Table \ref{tab:main_results} presents the quantitative results of our comprehensive evaluation comparing {\ours} against five distinct baselines: the unaligned Vanilla model, white-box steering (RepE), in-context safety (SysPrompt), generation-based intervention (SelfRefine), and the RLHF-aligned Official models. Overall, {\ours} demonstrates comprehensive superiority in balancing safety and response quality. With the exception of a marginal gap in safety scores on the \texttt{Llama-3-8B} backbone compared to the Official model, {\ours} consistently outperforms all baselines across safety, helpfulness, and naturalness metrics. Notably, on \texttt{Llama-3.1-8B}, {\ours} not only restores utility but also surpasses the Official model in safety (\textbf{9.34} vs. 9.18), validating its effectiveness in resolving the alignment trade-off.

A critical observation from the baselines is the severe ``alignment tax'' paid by the Official Instruct models. While they achieve high safety scores (e.g., \textbf{9.55} on Llama-3 and 9.18 on Llama-3.1), their helpfulness scores drop precipitously to 3.41 and 4.17, respectively. This discrepancy indicates a \textit{rigid refusal} behavior, where the model indiscriminately rejects sensitive queries. Moreover, the linguistic quality also degrades; for instance, the Naturalness score of the Official Llama-3 model drops to 6.71, falling even below the unaligned Vanilla baseline (7.04), which suggests that rigid safety filters compromise the conversational flow.
In contrast, {\ours} restores helpfulness to \textbf{7.89} (+4.48) and \textbf{7.79} (+3.62) on the Llama series while maintaining highly competitive safety levels. This confirms that our fine-grained guidance allows the model to navigate the ``safe boundary'' intelligently, generating nuanced responses that are both harmless and useful.

While other baselines such as RepE, SysPrompt, and SelfRefine attempt to improve upon the unaligned Vanilla model, they fall short of the balance achieved by {\ours}. For instance, RepE only provides a modest safety improvement over Vanilla (5.91 vs. 4.57 on Llama-3) and struggles to reach the high safety standards required. Similarly, SysPrompt improves safety to 8.40 but lags behind {\ours} in both safety (9.23) and helpfulness (7.89). Even SelfRefine, which achieves a respectable safety score of 9.03, suffers from a lower helpfulness score (6.65) and naturalness score (7.86) compared to our method. Furthermore, {\ours} consistently achieves the highest Naturalness scores across all backbones (e.g., \textbf{8.69} on Llama-3 vs. 7.74 for SysPrompt and 6.71 for Official). This indicates that our gradient-driven refinement preserves the linguistic fluency and original intent of the user, avoiding the mechanical or disjointed outputs often produced by inference-time interventions or rigid templates.

The robust performance on \texttt{Qwen2.5-7B} further underscores the generalizability of our framework. On this stronger backbone, which already exhibits high baseline performance (Vanilla Safety 9.74), {\ours} still achieves state-of-the-art results across all three metrics. It surpasses even the Official Instruct baseline in safety (\textbf{9.88} vs. 9.59), helpfulness (\textbf{8.03} vs. 7.58), and naturalness (\textbf{8.25} vs. 7.90). This demonstrates that {\ours} is not merely a patch for weaker models but a universal alignment solution that effectively circumvents rigid refusal behaviors while simultaneously enhancing safety, thereby fully preserving the inherent capabilities of advanced LLMs.

\subsection{Further Analysis}

\subsubsection{Ablation Studies}

To disentangle the contributions of our proposed components—specifically the necessity of the prompt refinement loop and the efficacy of the variational rejection distribution—we conduct an ablation study on the \texttt{Llama-3-8B-Abliterated} backbone. We compare the full framework against two degraded variants:

\begin{itemize}
    \item \textbf{w/o Refinement (\textsc{Lance-nr}):} A baseline configuration where the prompt optimization mechanism is entirely deactivated. This setting, representing the initial abliterated model mentioned earlier, serves as th lower bound to assess the raw capability of the model.

    \item \textbf{w/o Fine-Grained Guidance (\textsc{Lance-nf}):} A variant that ablates the label enhancement module to evaluate the necessity of rejection distribution. This variant relies on the refinement model (\texttt{Qwen2.5-1.5B-Instruct}) to judge rejection via a system prompt (e.g., ``You are an expert AI Safety...''). Consequently, it performs a coarse rewrite rather than a targeted refinement.
\end{itemize}

The results are presented in Table \ref{tab:ablation}. Comparing LANCE-NR with the other two methods, we observe that the static backbone yields a critically poor Safety score of 4.57. Introducing the optimization loop (even the coarse-grained one) dramatically boosts safety to 8.64. This confirms that dynamic iterative rewriting is a fundamental requirement for effectively aligning abliterated models.

\begin{table}[htbp]
\centering

\caption{Performance comparison to validate the effectiveness of the fine-grained refinement mechanism.}
\label{tab:ablation}


\setlength{\tabcolsep}{3pt}
\begin{tabular*}{\linewidth}{@{\extracolsep{\fill}} l ccc}
\toprule 
Variant & \bm{$S_{safe}$} ($\uparrow$) & \bm{$S_{help}$} ($\uparrow$) & \bm{$S_{nat}$} ($\uparrow$) \\ 
\midrule
\textsc{Ours-nr} & 4.57 & 5.59 & 7.04 \\
\textsc{Ours-nf} & 8.64 & 7.69 & 8.37 \\

\textsc{Ours}  & \textbf{9.23} & \textbf{7.89} & \textbf{8.69} \\
\bottomrule
\end{tabular*}
\end{table}

The key comparison lies between LANCE-NF and our Full Method {\ours}. While the coarse-grained variant significantly improves safety, it lags behind our method across all metrics. Our method improves $\mathcal{S}_{safe}$ from 8.64 to \textbf{9.23}. This indicates that the continuous rejection distribution $\bm{d}$ via label enhancement captures subtle, latent risks that the LLM's intrinsic binary judgment fails to detect. Crucially, our method achieves higher Helpfulness and Naturalness. The coarse-grained baseline relies on binary triggers (safe/unsafe), which often leads to abrupt or heavy-handed rewrites when the safety boundary is fuzzy. In contrast, our fine-grained rejection distribution allows for adaptive gradients—applying ``Minor'' or ``Mild'' edits to low-risk prompts. This preserves the original semantic structure and fluency, proving that the continuous rejection distribution is essential for breaking the rigid refusal trade-off.

\subsubsection{Experimental Validation of Label Enhancement}

To validate the efficacy of label enhancement in mitigating rigid refusal, we conducted an independent evaluation of the core label-augmented module on the XSTest over-rejection dataset. As shown in  Table \ref{tab:label}, we systematically scanned and quantitatively analyzed the module’s performance across various risk thresholds ($\tau$) on the dataset. For safe yet potentially edge-case queries (Safe group), we focused on the False Positive Rate (FPR)—the proportion of benign queries misclassified as risky and erroneously rejected. Conversely, for genuine harmful instructions (Harmful group), we assessed Detection Rate, measuring the model’s ability to correctly identify and intercept malicious inputs. The results highlight the module’s balanced trade-off between minimizing over-rejections and ensuring safety.
\begin{table}[htbp]
    \centering
    \caption{Performance of the label enhancement module on the XSTest dataset.}
    \label{tab:label}
    \begin{tabular}{@{}lcc@{}}
        \toprule
        \textbf{Risk ($\tau$)}& \textbf{FPR (Safe) $\downarrow$} & \textbf{Detection (Harmful) $\uparrow$} \\
        \midrule
        0.3 & 21.67\%& 96.38\%\\
        0.5 & 13.71\%& 92.58\%\\
        0.7 & 8.81\%& 90.64\%\\
        \bottomrule
    \end{tabular}
\end{table}

\subsubsection{Comparative with Commercial Models }

To further contextualize the efficacy of our proposed framework, we challenge our lightweight approach against state-of-the-art commercial large models, including \texttt{Kimi-k2}, \texttt{Qwen3-Max}, \texttt{Gemini2.5-Pro} and \texttt{GPT-4o}. For a fair comparison, we report the averaged performance of our method across the three abliterated backbones. 

As shown in Table \ref{tab:commercial_comparison}, our comparative analysis exposes a critical limitation in various top-tier commercial models: their reliance on a ``safety-by-silence'' strategy. While \texttt{Gemini2.5-Pro} and \texttt{GPT-4o} achieve exceptional safety scores, they suffer from critically low Helpfulness scores (2.84 and 3.58, respectively). This quantitative gap points to severe \textit{rigid refusal} behaviors, where models indiscriminately block borderline queries with mechanical refusal templates (e.g., ``Sorry, I cannot fulfill this request'' or ``I can't help with that'').
\begin{table}[h]
    \centering
    \caption{Comparison with Commercial Large Models. }
    \label{tab:commercial_comparison}
    \begin{tabular}{lccc}
        \toprule
        Model& \bm{$S_{safe}$}($\uparrow$)& \bm{$S_{help}$} ($\uparrow$)& \bm{$S_{nat}$} ($\uparrow$)\\
        \midrule
        Kimi-k2& 9.68 & 7.35 & 8.13 \\
        Qwen3-max& \textbf{9.89}& 7.81& 8.35\\
        Gemini2.5-Pro& 9.33& 2.84& 4.12\\
 GPT-4o& 9.84& 3.58&6.64\\
 \midrule
        Ours& 9.55& \textbf{7.88}& \textbf{8.53}\\ \bottomrule
    \end{tabular}
    
\end{table}

In sharp contrast, our method achieves a Helpfulness score of \textbf{7.88}. Our framework leverages fine-grained refinement guidance to optimize the prompt, a process that eliminates rigid rejections and transforms what would otherwise be unsolvable rejections into constructive and safe responses.
Despite relying on significantly smaller open-source backbones (7B/8B parameters) and a lightweight refinement model (1.5B), our framework maintains a competitive Safety score of \textbf{9.55}. This result proves that scale is not the sole prerequisite for effective alignment. By replacing coarse rejection filters with our disentangled risk detection mechanism, {\ours} delivers a superior safety-utility trade-off.

\section{Conclusion}

In this paper, we introduce {\ours}, a novel framework designed to dismantle the pervasive ``rigid refusal'' paradigm in Large Language Model (LLM) safety alignment. By leveraging a disentangled variational inference mechanism, we successfully isolate safety constraints (via a Beta-distributed rejection latent) from linguistic content (via a Gaussian semantic latent), which transforms coarse risk labels into a dense, interpretable rejection distribution. Empowered by this precise diagnostic, our gradient-driven refinement model performs adaptive prompt refinement, surgically neutralizing harmful intentions while preserving the user's original semantic style. Ultimately, {\ours} demonstrates that rigid refusal is not an inevitability of safety alignment, offering a robust solution that restores model utility and naturalness while maintaining rigorous safety compliance.

\section*{Impact Statement}

This paper presents work whose goal is to advance the field of Machine
Learning. There are many potential societal consequences of our work, none
which we feel must be specifically highlighted here.


\bibliography{example_paper}
\bibliographystyle{icml2026}

\end{document}